\documentclass[11pt,a4paper]{article}
\usepackage[hyperref]{acl2021}
\usepackage{times}
\usepackage{latexsym}
\usepackage{xcolor}
\usepackage{amsmath}
\usepackage{algorithm,algpseudocode}
\usepackage{amsfonts}
\usepackage{amssymb}
\usepackage{xspace}
\usepackage{graphicx}
\usepackage{bbm}
\usepackage{dsfont}
\usepackage{comment}

\usepackage{microtype}

\aclfinalcopy 

\newcommand{\system}{PR-VIST\xspace}

\title{Plot and Rework: Modeling Storylines for Visual Storytelling}

\author{Chi-Yang Hsu$^{1,3}$\thanks{* denotes equal contribution}, Yun-Wei Chu$^{2 *}$, 
Ting-Hao (Kenneth) Huang$^1$, 
Lun-Wei Ku$^{3}$ \\
    Pennsylvania State University ${^1}$, Purdue University ${^2}$, \\ Institute of Information Science, Academia Sinica${^3}$ \\
    {\tt \{cxh5437, txh710\}@psu.edu}\\
    {\tt \{chu198\}@purdue.edu}\\
  {\tt \{lwku\}@iis.sinica.edu.tw}
  }

\date{}

\begin{document}
\maketitle
\begin{abstract}

Writing a coherent and engaging story is not easy.
Creative writers use their knowledge and worldview to put disjointed elements together to form a coherent storyline, and work and rework iteratively toward perfection.
Automated visual storytelling (VIST) models, however, 
make poor use of  
external knowledge and iterative generation when attempting to create stories.  
This paper introduces \textit{\textbf{\system}}, a framework that 
represents the input image sequence as a story graph
in which it finds the best path to form a storyline.
\system then takes this path and learns to generate the final story via a re-evaluating training process.
This framework produces stories that are superior in terms of diversity, coherence, and
humanness, per both automatic and human evaluations.
An ablation study shows that both plotting and reworking contribute to the
model's superiority.





\end{abstract}

\section{Introduction}

\begin{figure*}[t]
    \centering
    \includegraphics[width=\linewidth]{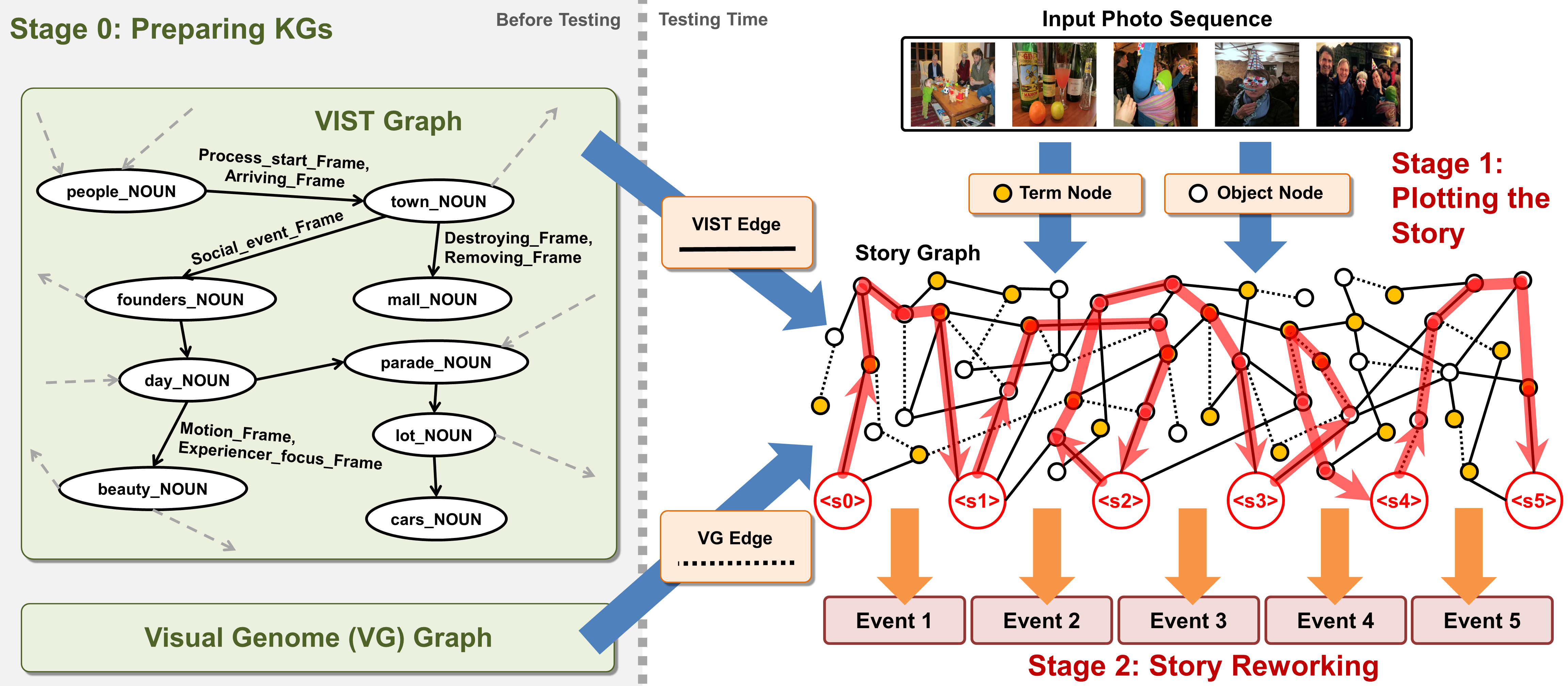}
	\caption{Overview of {\system}. In \textbf{Stage 1 (Story Plotting)}, {\system} first constructs a graph that captures the relations between all the elements in the input image sequence and finds the optimal path in the graph that forms the best storyline. In \textbf{Stage 2 (Story Reworking)}, {\system} uses the found path to generate the story. 
	PR-VIST uses a story generator and a story evaluator to realize the ``rework'' process.
	In \textbf{Stage 0 (Preparation)}, a set of knowledge graphs that encode relations between elements should be prepared for the uses in Stage 1.}
	\label{fig:flowchart}
\end{figure*}

Writing a story is a complicated task.
Human writers use their knowledge to tie all the disjointed elements, such as
people, items, actions, or locations, together to form a coherent storyline.
Writers also re-evaluate their work constantly during the writing process, and
sometimes even alter their writing goals in the middle of a draft.
\citet{flower1981cognitive} characterize a solo writer's cognitive process as
a series of components in which the writer's own knowledge is described as the
long-term memory,
and the planning, translating, and reviewing steps can occur in a
recursive, interconnected manner.
These creative steps are essential to human writing.
However, automated visual storytelling (VIST) models that
compose stories given five images~\cite{huang2016visual} 
do not make extensive use of human knowledge to tie the elements together,
nor do they use human insight to evaluate the outputs and guide the generation process.

As for linking elements, most works generate visual stories in an
end-to-end fashion~\cite{huang2016visual,kim2018glac}, treating the task as a
straightforward extension of image captioning.
Recent works have begun to use relations between entities to improve visual
storytelling, but often narrow in a particular subset of relations, such
as 
relations between elements within the same image~\cite{pengchengKnowledge2019},
relations between two adjacent images~\cite{hsu2020knowledge}, or 
relations between scenes~\cite{Wang2020StorytellingFA}.
The full potential of rich real-world knowledge and intra-image
relations have yet to be fully utilized.
As for re-evaluation, recent work uses reward
systems~\cite{DBLP:journals/corr/abs-1804-09160,hu2019makes} or estimated topic
coherence~\cite{wang2019consistent} to automatically assess the output story and guide the
generation process.
However, these approaches are often optimized towards predefined aspects such as
image relevancy or topic coherence, which do not necessarily lead to 
engaging stories from a human perspective.
In the cognitive process of human writing, the writer's judgment is critical,
and visual storytelling models could benefit by considering human ratings.

This paper introduces \textit{\textbf{\system}}, a novel visual storytelling
framework that constructs a graph and captures the relations between all
the elements in the input image sequence,
finds the optimal path in the graph that forms the best storyline,
and uses this path to generate the story.
An overview of \system is shown in Figure~\ref{fig:flowchart}.

\begin{itemize}
	 \item \textbf{Stage 1 (Story Plotting):} \system first constructs a story
	 graph for the image sequence by extracting various elements (i.e.,
	 term nodes, object nodes) from all the images and linking these elements 
	 using external knowledge (i.e., VIST graph, VG graph). \system then
	 finds the best path in the story graph as the storyline and passes it
	 to Stage~2.

\item \textbf{Stage 2 (Story Reworking):} \system uses a story generator and a
story evaluator to realize the reworking process: the generator takes the
storyline produced in Stage~1 as the input to generate the story and
backpropagates with an evaluator-augmented loss function. 
The evaluator, a discriminator model trained on human rating score data to
classify good and bad stories, outputs a story quality score and modifies the
loss. 
After a few optimization epochs, the generator eventually learns to generate
stories that 
reflect human preferences.

\end{itemize}

In \textbf{Stage 0 (Preparation)}, a set of knowledge graphs that encode
relations between elements are prepared for use in Stage~1.
In this work, we prepare two knowledge graphs: a VIST graph and a visual
genome (VG) graph. 
We construct the VIST graph based on the VIST dataset, representing 
in-domain knowledge; the VG graph is an existing resource~\cite{Krishna_2017},
representing generic knowledge.
Note that as the \system framework is generic, it can use any knowledge graphs as needed.

Automatic and human evaluations show that \system produces visual stories
that are more diverse, coherent, and human-like.
We also conduct an ablation study to show that both story plotting (Stage~1)
and reworking (Stage~2) contribute positively to the model's superiority.
We believe this work also shows the potential of drawing inspiration from
human cognitive processes and behavior to improve text generation
technology.

\section{Related Work} 


\paragraph{Visual Storytelling}
Researchers have been trying to advance the visual storytelling task since it was introduced by~\citet{huang2016visual}.
Some work modifies end-to-end recurrent models for better story generation~\cite{hsu2018using,gonzalez2018contextualize,kim2018glac,huang2019hierarchically,
Jung2020HideandTellLT}, and some use adversarial training to generate
more diverse
stories~\citep{chen2017multimodal,wang2018show,DBLP:journals/corr/abs-1804-09160,
hu2019makes}.
These methods produce legitimate stories and easier to implement because they relies only on one dataset.
However, the generated stories can sometimes be monotonous and repetitive.

\paragraph{Leveraging External Resources for VIST}
Another set of work leverages external resources and knowledge to enrich the generated visual stories. 
For example, \citet{pengchengKnowledge2019} apply ConceptNet~\citep{liu2004conceptnet} and self-attention for create commonsense-augmented image features;
\citet{Wang2020StorytellingFA} use graph convolution networks on scene graphs~\cite{Johnson2018ImageGF} to associate objects across images; and
KG-Story~\cite{hsu2020knowledge} is a three-stage VIST framework that uses Visual Genome~\cite{Krishna_2017} to produce knowledge-enriched visual stories.


\paragraph{Editing or Optimizing Visual Stories}
A few prior work tries to post-edit visual stories or optimize the story content toward specific goals.
VIST-Edit is an automatic post-editing model that learns from an pre- and post-edited parallel corpus to edit machine-generated visual stories~\cite{Hsu2019VisualSP}.
While VIST-Edit is useful, it requires parallel training data, which is often unavailable.
\citet{hu2019makes} use a reward function to optimize the generated stories toward three aspects; \citet{10.1145/3323873.3325050} customize the emotions of visual stories.
These methods use automatic metrics to optimize visual stories toward specific goals;
our work, on the other hand, leverages the human evaluation data to guide the generation process.

\paragraph{Story Plotting in Story Generation}
Research in automatic story generation has demonstrated the
effectiveness of story plotting~\cite{DBLP:journals/corr/abs-1811-05701,
Fan2019StrategiesFS}, which typically involves organizing the ``ingredients'' into a well-organized sequence of events. 
Nevertheless, none of the studies applied story plotting for visual stories.


\section{Stage 0: Preparation} 

To prepare for story plotting, we collect information from the images
and knowledge from the knowledge graphs. 

\subsection {Story Element Extraction} 
To extract information from the images, two extraction methods are used to
extract image-oriented and story-oriented story elements: \emph{objects} and
\emph{terms}, respectively representing image and story intuition.

\paragraph{Objects} These can be detected by current object detection
models, for which we use a pre-trained object detection
model---Faster-RCNN~\cite{ren2015faster}. To ensure the detected objects' reliability, only those objects with the top five confidence scores are used
in each image.
\paragraph{Terms} These are story-like nouns such as events, time, and
locations, which current object detection models are unable to extract.
Therefore, we further use a Transformer-GRU~\cite{hsu2020knowledge} to predict
story-like terms. For each image and story pair, we use image objects as the
input and the nouns in the corresponding human-written story as the
ground truth. The Transformer-GRU learns to convert objects to 
  nouns          commonly used in     stories.  


\subsection{Knowledge Graph Preparation}
To collect interactive relations between nouns,  
we prepare Visual Genome graph  \(\mathcal{G}_{\mathit{vg}}\) and VIST graph
\(\mathcal{G}_{\mathit{vist}}\). 
These graphs contain interlinked real-world objects and terms, displaying
visual and storytelling interaction.
Table~\ref{tb:kg_table} summarizes the statistic of each graph.

\paragraph{Visual Genome Graph}
{\(\mathcal{G}_{\mathit{vg}}\)} describes pairwise relationships between
objects in an image, describing visual interactions. 
No prepositional relations are included; only verb relations are
preserved. All relations are converted into semantic verb frames using
Open-SESAME~\cite{swayamdipta17open}, in which the semantic frames were
pre-defined in FrameNet~\cite{baker1998berkeley}.   
\paragraph{VIST Graph}
we propose \(\mathcal{G}_{\mathit{vist}}\) to collect the storytelling
interactions. We develop this novel story knowledge graph
by converting references in the VIST training and validation
datasets~\cite{huang2016visual} to
graphical data. 
\begin{figure}[t]
    \centering
    \includegraphics[width=0.95\linewidth ]{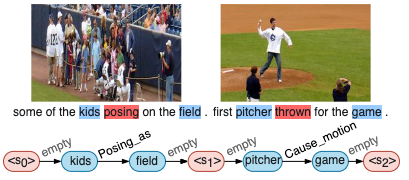}
	 \caption{From the stories in VIST training data, a story is transformed into a
	 golden storyline following the human reading direction, as part of the
	 VIST graph.}
    \label{fig:vist-graph}
\end{figure}
Following the reading direction, in each reference, we extract nouns and
semantic verb frames using SpaCy\footnote{SpaCy: https://spacy.io/} and
Open-SESAME to obtain noun--verb--noun (NVN) tuples.
Using nouns and semantic verb frames as nodes and edges, these are collectively
assembled into a golden storyline. 
For example, for ``first pitcher thrown for the game'' in
Figure~\ref{fig:vist-graph}, we extract \emph{pitcher}, \emph{game}, and
\emph{Cause\_motion}, which is a semantic verb frame for \emph{thrown}, as a
NVN tuple.
Additionally, we include a noun token \texttt{$<$s{\scriptsize{i}}$>$} as the
transition point to the next sentence or termination point of a story, and a
verb frame token \texttt{empty\_frame} to interlink two nouns when a semantic
frame is absent. To conclude, all of the golden storylines are assembled into
\(\mathcal{G}_{\mathit{vist}}\).

\begin{table}[]
\begin{center}

\begin{tabular}{lccc}
\textbf{} & Nodes & Relations & Links\\ \hline
\(\mathcal{G}_{\mathit{vg}}\)         & 3,323        & 564             & 22.31                 \\
\(\mathcal{G}_{\mathit{vist}}\)      & 2,048        & 531             & 11.75                 \\
\(\mathcal{G}_{\mathit{vg+vist}}\)    & 4,158        & 880             & 22.78                 \\ \hline
\end{tabular}

\caption{The statistics of knowledge graphs. The table shows the number of distinct nodes and relations in each graph. It also shows the average link per node. Note that the nodes and relations from \(\mathcal{G}_{\mathit{vg}}\)  and \(\mathcal{G}_{\mathit{vist}}\) have overlaps. 
}
\label{tb:kg_table}
\end{center}
\end{table}

\section{Stage 1: Story Plotting}
\subsection{Storyline Predictor}

In Stage~1, \system uses a storyline predictor to find what it deems the best path in the
story graph as the storyline and then pass this to Stage~2.
For the storyline predictor, we use UHop~\cite{chen2019uhop}, a non-exhaustive relation extraction
framework.
A single hop is defined as searching from one entity to another entity by a
single relation.
UHop performs multiple single-hop classifications consecutively in the graph to
  find the path representing the storyline,  
that is, a path that consists of a sequence of nouns and
verb frames. 

\begin{figure}[t]
    \centering
    \includegraphics[width=0.95\linewidth]{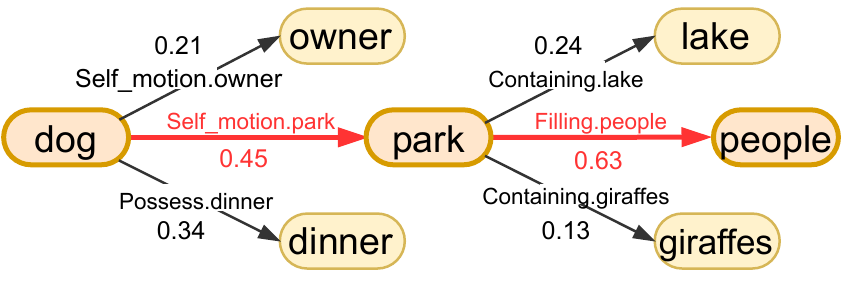}
	 \caption{Storyline pathfinding process. All entities are from \emph{object} or \emph{term} list, and all relations are in [verb.noun] format, which the verbs are verb frames from knowledge graphs and the nouns are the tail entities. 
     The single-hop classification begins
	 with \texttt{dog}. 
	 The storyline predictor is given three candidate relations. The framework selects the highest score
	 relation and move on to the next entity \texttt{park}. Then, the single-hop
	 classification repeats. 
    } 
    \label{fig:UHop}
\end{figure}

Single-hop classification can described as Equation~\ref{UHop_equation} and
Figure~\ref{fig:UHop}.
In step~\(i\), at the current head entity \(h_{i}\), the model is given a list of
candidate relations \(r_{i} \in R_{i}\) and tail entities \(t_{i} \in T_{i}\). Each
\(r_{i}\) is in [verb.\(t_{i}\)] or [verb.noun] format, containing 
  information for both the verb frame and the tail noun entity.   
The scoring model \(F\) is given \emph{objects} and predicted relations
\(r_{1},...,r_{i-1}\) as input. The model predicts a score for each \(r_{i}\)
and selects the highest verb-noun pair \(r_{i}\) from \(Q\):
{
\begin{equation} 
    \label{UHop_equation}
    r_{i} = \mathop{\arg\max}_{q \in Q}  F(\mathit{objects}, r_{1},...,r_{i-1}).
\end{equation}
}
\paragraph{Training} UHop learns to find a path for the storyline from the golden
storyline. The training procedure starts with an initial noun token entity
\texttt{$<$s0$>$} in the golden storyline for single-hop classification,
where \(h_{1}=\) \texttt{$<$s0$>$}. It learns to select the correct relation
\(r_{i}\) from a list of candidate relations \(R_{i}\) in
\(\mathcal{G}_{\mathit{vg}}\) and \(\mathcal{G}_{\mathit{vist}}\). Then, it
calculates the error to the noun and verb frame in the golden storyline for
backpropagation. In the next hop, the framework proceeds to the next noun in the
golden storyline and repeats the single-hop classification.

\paragraph{Testing} 
In \system's testing step, for each story, five images are transformed into a
story graph \(\mathcal{G}_{\mathit{story}}\). 
As demonstrated in Figure~\ref{fig:flowchart}, we first extract the
\emph{object} and \emph{term} story elements for each story, and then link
these together using the verb frames in \(\mathcal{G}_{vist}\) and
\(\mathcal{G}_{vg}\) as edges. This yields a
well-defined graph presenting a
comprehensive view of five images for each
story---\(\mathcal{G}_{\mathit{story}}\). Next, a trained UHop finds a
storyline in  \(\mathcal{G}_{\mathit{story}}\), where all entities are only the
\emph{objects} and \emph{terms} from the given five images. The framework starts with
\texttt{$<$s0$>$} to perform single-hop classification, where \(h_{1}=\)
\texttt{$<$s0$>$}. Unlike training, it only selects \(r_{i}\) from \(R_{i}\)
listed in \(\mathcal{G}_{\mathit{story}}\). In the next hop, 
the previous predicted entities are used as the start entity: \(h_{0}=t_{i-1}\). It then
continues to hop from entity to entity until it reaches the next token
\texttt{$<$s1$>$}. The path from \texttt{$<$s{\scriptsize{i-1}}$>$} to
\texttt{$<$s{\scriptsize{i}}$>$} is called an event $e_{i}$. The path search
from \texttt{$<$s{\scriptsize{i}}$>$} to the next token
\texttt{$<$s{\scriptsize{i+1}}$>$} continues until the search is terminated by
the termination decision described in UHop. Eventually, the model finds a storyline
of arbitrary length $L$, that is, a storyline that contains any number of events:
$e_{1},...,e_{L}$.

\subsection{Implementation Details}



HR-BiLSTM~\cite{Yu_2018} is adopted as the scoring model $F$, in which
\emph{objects} are converted to word embeddings via
GloVe~\cite{pennington2014glove} as {$E(\mathit{object})$}. 
All relation embeddings $\mathit{E}(r)$
are decomposed into graphical embedding $\mathit{E}_{\mathit{graph}}$ and textual
embedding $\mathit{E}_{\mathit{text}}$. 
$\mathit{E}_{\mathit{graph}}$ transforms a verb frame $v$
and a tail entity's image position $p^{t}$ into a one-hot vector, denoting the
graphical and image positional information. 
$\mathit{E}_{\mathit{text}}$ is composed of the verb frame and tail entity word embedding.
Then, $\mathit{E}_{\mathit{graph}}$ and $\mathit{E}_{\mathit{text}}$ are concatenated into a unified
representation $\mathit{E}(r)$. 
We formulate the representation of relation~$r$ as
{
\begin{equation}
    \begin{split}
    \mathit{E}_{\mathit{graph}}(r) = [\mathbbm{1}(v) ; \mathbbm{1}(p^{t})], \\
    \mathit{E}_{\mathit{text}}(r) = [E_{w}(f); E_{w}(t)], \\
    \mathit{E}(r) = [\mathit{E}_{\mathit{graph}}(r) ; \mathit{E}_{\mathit{text}}(r)],
    \end{split}
\end{equation}
}%
where $\mathbbm{1}(\cdot)$ returns the one-hot representation, 
$E_{w}(\cdot)$ returns the word embeddings via
GloVE, and [;] denotes concatenation.

A verb frame and tail entity are combined into $r_{i}$ due to relational
ambiguity issues among candidate relations.
Using Figure~\ref{fig:UHop} as an example, given a head entity \emph{dog}, candidates
\emph{self\_motion.park} and \emph{self\_motion.owner} represent different semantic
meanings when tail entities \emph{park} and \emph{owner} are included. However,
excluding tail entities results in identical relation candidates
\emph{self\_motion} and thus ambiguity between two different candidates.









\section{Stage 2: Story Reworking}
In story reworking, the framework consists of two components: the story
generator and the story evaluator. The story generator generates a story
according to the storyline, and the story evaluator---a discriminator trained on
the MTurk human ranking data to classify good and bad stories---outputs a story
quality score and modifies the loss functions.



\subsection{Story Generator} 
A storyline consists of a set of events  $e_{1}...e_{L}$ that are input to the
story generator, which is based on the Transformer~\cite{vaswani2017attention}. 
Unlike most VIST models, the story generator is dynamic: the number of output
sentences depends on the number of events. 
To manage a diverse number of events, the Transformer is designed as a sentence
generator that iteratively generates one sentence per event until it generates
$L$ sentences. 
For each step~$i$, event~$e_{i}$ and the previous
predicted sentence $y_{i-1}
$
are used to predict the next sentence $y_{i}$.
After $L$~steps, the story generator outputs a story $s = y_{1},...y_{L}$.




\subsection{Story Evaluator}
Most VIST works use human evaluations to examine their work's quality via crowdsourcing, comparing their generated stories with the baseline
stories. In this paper, we use the first- and last-place stories in the MTurk human
ranking data as positive and negative samples. The story evaluator, a
discriminator trained on the MTurk human ranking data, learns to distinguish
positive and negative samples. It outputs a score for each story, and the
scores are converted into rewards, as shown below:
%
{
\begin{equation} 
    p_{\mathit{LM}}(u|s) = {\rm softmax}({\rm tanh}(\textbf{W} \mathit{LM}(s)) + \textbf{b} ),
\label{scoring_stories}
\end{equation}
}%
{
\begin{equation} 
    \mathcal{R}(s) = - p_{\mathit{LM}}(u|s) + c,
\label{reward_scoring}
\end{equation}
}%
where $\mathit{LM}(\cdot)$ is a GRU language model~\cite{cho2014learning}, $u=1$
indicates story~$s$ is a positive sample, and $u=0$ indicates $s$ is a negative
sample. 
Language model $p_{\mathit{LM}}(\cdot)$ returns a score from 0 to 1 to reflect story
quality. 
The story evaluator $\mathcal{R}(\cdot)$ returns a reward, an inverse of
$p_{\mathit{LM}}(\cdot)$ with coefficient $c = 1.5$. The reward later
manipulates the loss, optimizing toward human preference. Note that the story
evaluator is pre-trained. 


\subsection{Optimization with Story Evaluator}
\begin{figure}[t]
    \centering
    \includegraphics[width=0.95\linewidth ]{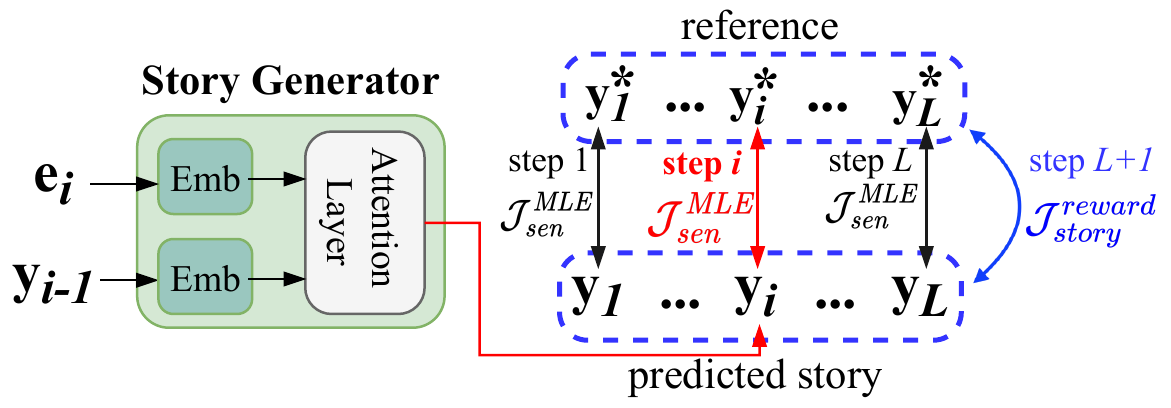}
	 \caption{Optimization flowchart for story generator. For steps 1 to $L$, the
	 model is optimized using sentence-level loss. In step $L+1$, all sentences are
	 generated, and the model is 
 	 optimized using story-level loss.   
	 }
    \label{fig:loss}
\end{figure}

For optimization, the story generator uses sentence-level and story-level loss functions.
Given reference $y^*_{1},...,y^*_{L}$ and predicted story $y_{1},...,y_{L}$, in
the maximum likelihood estimation (MLE) optimization process, in each step from
1~to~$L$, the model predicts a sentence $y_{i}$ to calculate the loss between
$y_{i}$ and $y^*_{i}$ and then backpropagates, as shown in Figure~\ref{fig:loss}.
After predicting $L$ sentences, in story-level optimization, the model predicts
$y_{1},...,y_{L}$ to calculate the negative log-likelihood to the reference
$y^*_{1},...,y^*_{L}$ and then backpropagates. 
The sentence-level and story-level optimization by MLE on dataset
$\mathcal{D}$ are formulated as 
{
\begin{equation} 
\label{sen_equation}
    J^{\mathit{MLE}}_{\mathit{sen}}(\theta, \mathcal{D}) = 
    \sum_{Y \in \mathcal{D}} - \log p_{\theta}(y_{i}| e_{i}, y_{i-1}), 
\end{equation}
}%
{
\begin{equation} 
\label{story_equation}
    J^{\mathit{MLE}}_{\mathit{story}}(\theta, \mathcal{D}) = 
    \sum_{Y \in \mathcal{D}} \frac{1}{L}\sum_{i=1}^{L} - \log p_{\theta}(y_{i}'| e_{i}, y_{i-1}'),
\end{equation}
}%
where $e_{i}$ 
and $y_{i}$ denote the $i$-th event 
and the sentence respectively, 
$y_{i}'$ represents the updated sentence after sentence-level optimization, 
and $\theta$ represents the story generation model parameters, 
which are updated using Adam~\cite{kingma2014adam}. 

After training for 30~epochs \footnote{The generation model converges at around 20 epochs in our experiment, and we give it extra 10 epochs for precautions.}, the story evaluater begins to manipulate the
story-level loss. Inspired by reinforcement learning~\cite{10.1007/BF00992696},
which utilizes rewards to guide the training process, we use the story evaluator
$\mathcal{R}(\cdot)$ \footnote{The pre-trained $\mathit{LM}$'s weights are frozen to stabilize the training. } to encourage the generation model to focus on stories
preferred by humans. 
The reward directly multiplies the story-level loss as

%
{
\begin{small}
\begin{equation} 
    \label{reward_equation}
    J^{\mathit{reward}}_{\mathit{story}}(\theta, \mathcal{D}) = 
    \begin{cases}
    J^{\mathit{MLE}}_{\mathit{story}} & \text{if epoch $\leq$ 30}\\
    \mathcal{R}(s)J^{\mathit{MLE}}_{\mathit{story}} & \text{if 30 $<$ epoch $\leq$ 60}
    \end{cases}.
\end{equation}
\end{small}
}%

\section{Experimental Results}

\subsection{Data Setups}
We used four datasets in this paper: the VIST dataset, Visual Genome, ROCStories,
and MTurk human ranking data.
The VIST dataset and Visual Genome are used to construct the knowledge graphs, and 
ROCStories~\cite{Mostafazadeh2016ACA} is a large quantity of pure textual stories used for pre-training
the story generator.
The VIST dataset is also used in story plotting to train the storyline predictor
and in story reworking to fine-tune the story generator.
Notably, we also collected MTurk human ranking data to train the story evaluator.
We used the ranking results from KG-Story\footnote{Data obtained from the
authors of KG-Story.}~\cite{hsu2020knowledge}.
This data contains two experiments, each of which  ontains 500 distinct
photo sequences. A photo sequence contains a set of machine-generated stories
ranked by 5 MTurk workers. Thus we have 5000 rankings from MTurk workers.
Specifically, MTurk workers were asked to rank AREL~\cite{wang2018show},
KG-Story, two KG-Story ablation models, and reference stories, using three
different model settings in each experiment.
We selected the rank-1 and rank-5 stories as positive and negative samples.

\subsection{Baselines}
We used several competitive baseline models. 
\textbf{AREL}~\cite{wang2018show} and \textbf{GLAC}~\cite{kim2018glac} are
end-to-end models with 
reinforcement learning and
global-location attention mechanisms that achieved top ranks in the VIST
Challenge~\cite{mitchell2018proceedings}.
\textbf{KG-Story}~\cite{hsu2020knowledge}, the current state-of-the-art framework, 
utilizes a knowledge graph to 
enrich story contents
and generates stories using
Transformer.





\subsection{Evaluation Methods}

Per the literature~\cite{wang2018show}, human evaluation is the
most reliable way to evaluate the quality of visual stories; automatic
metrics often do not align faithfully to human
judgment~\cite{Hsu2019VisualSP}.
Therefore, in this paper, we prioritize human evaluation over automatic evaluations.

\paragraph{Human Evaluation: Ranking Stories and Filling a Questionnaire}
\label{sec:human-eval-method}
We recruited crowd workers from Amazon Mechanical Turk (MTurk) to assess the quality of the generated stories. 
For each experiment, we randomly selected 250 stories, each of which was
evaluated by five different workers in the US.
The experiment includes a
comparison study with three baseline models and three ablation studies, and
each annotator was compensated \$0.10 for each experiment. 
Workers were asked to rank the stories ({\em e.g.}, ours and those of the
baseline/ablation models) based on their overall quality. 

In addition, the user interface also provides a questionnaire to collect
in-depth feedback from MTurk workers. The questions include ``What do you like
about the best stories'' and ``What do you dislike about the worst stories''
for workers to select aspects that affect overall story quality.  
These aspects are provided by~\citet{huang2016visual}: they include focus,
coherence, shareability, humanness, grounding, and detail.
We calculated the average rank and the majority rank among five workers for each
story, as well as total votes for each model's best and worst aspects.

\paragraph{Non-Classic Automatic Evaluation: BLEURT, voc-d, and MLTD}
Many VIST studies have shown that classic automatic
evaluation scores like BLEU and METEOR correlate poorly with human
judgment~\cite{hsu2020knowledge, hu2019makes, Wang2020StorytellingFA, Li_2020,
pengchengKnowledge2019, Hsu2019VisualSP, wang2018show,
modi-parde-2019-steep}.
These n-gram matching metrics fail to account for the semantic similarity to the
reference stories and lexical richness in the generated stories. 

Therefore, we adopted BLEURT~\cite{sellam2020bleurt}, a state-of-the-art BERT-based
evaluation metric, to further correlate generated stories and reference
stories based on their semantic meaning. We also adopted lexical
diversity metrics voc-d and MLTD~\cite{Lexical} to quantify story lexical
richness. 
Several works have shown that lexical diversity is positive correlated to story
quality~\cite{Liu_2019_ICCV, dai2017towards}. 
\begin{figure}[t]
    \centering
    \includegraphics[width=\linewidth ]{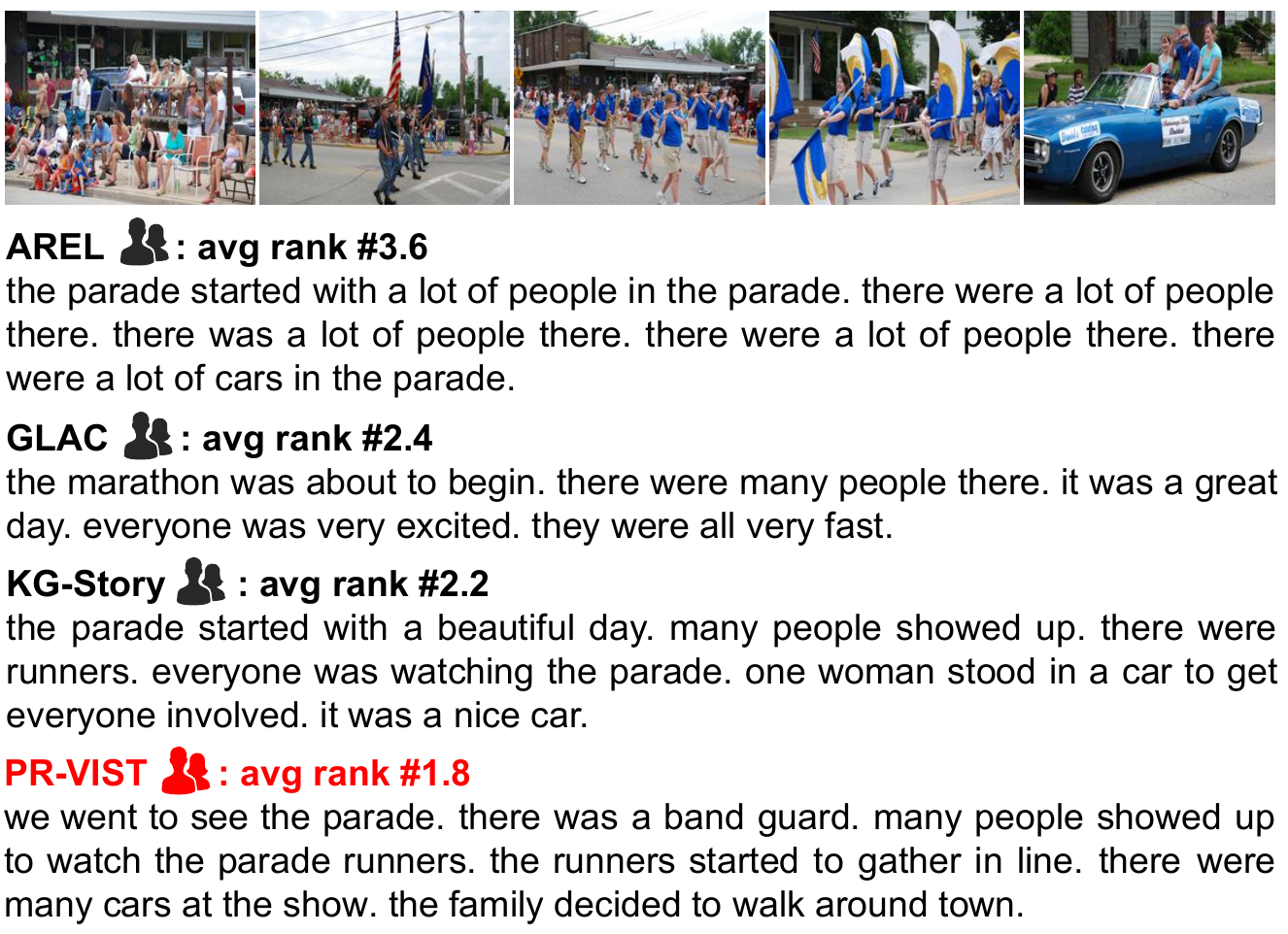}
	 \caption{Generated stories for \system and baseline models. MTurk workers
	 were asked to rank the stories. 
	 }     
    \label{fig:baseline_comparison}
\end{figure}


\subsection{Results}
In our experiments, the stories generated by {\system} have an average of 5.96 sentences.
57.3\% of these stories contain at least one event (sentence) that uses story elements extracted from two (or more) images, showing \system's ability to utilize intra-image entities.

\paragraph{Human Evaluation}

We asked MTurk workers to
rank four stories: those of \system, the three baseline models,
and the state-of-the-art KG-story.
Table~\ref{tb:human_ranking} shows the results.
\system outperforms other models in average ranking: it outranks AREL by 0.24
and KG-Story by 0.16. 
As for the percentage of 1st-rank stories, \system produces 12.0\% more than
AREL and 7.5\% more than KG-Story.
Figure~\ref{fig:baseline_comparison} shows a representative example.
Compared with end-to-end models ({\em i.e.}, AREL and GLAC), graph-based
methods ({\em i.e.}, KG-Story and \system) generate more diverse stories.
Compared with KG-Story, whose sentences are relatively simple and plain, generating sentences such
as ``Many people showed up'', our model reuses entities such as ``parade''
in the first sentence and associates relations with other entities, {\em e.g.},
``people'' and ``runners'', to compose ``many people showed up to watch
the parade runners''. 

\begin{figure}[t]
    \centering
    \includegraphics[width=\linewidth ]{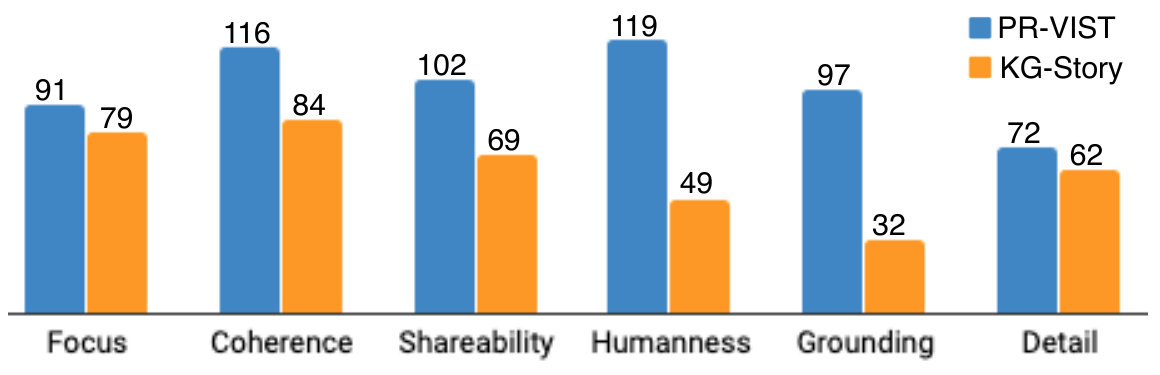}
	 \caption{Aspect-wise votes for {\system} and KG-Story's first-place 
	 stories collected via the questionnaire (see Section~\ref{sec:human-eval-method}). \system outperforms drastically in coherence, humanness,
	 and grounding.
} 
\label{fig:ours_kgstory_bar2}
\end{figure}

Moreover, Figure~\ref{fig:ours_kgstory_bar2} shows the questionnaire (see Section~\ref{sec:human-eval-method}) result for
the best-ranked stories.
For {\system} and KG-Story's best-ranked stories, the {\system} story count is
significantly higher in all aspects; specifically, coherence, shareability, and
humanness are higher than other categories.

\begin{table}
\begin{center}
\small
\scalebox{0.79}{

\begin{tabular}{lcccccc}
\hline
Method & 1st & 2nd & 3rd & 4th & Avg & Major \\
\hline\hline
AREL  & 20.6\% (258) & 26.8\% & 27.2\% & 25.4\% & 2.57 & 2.56 \\
GLAC  & 21.7\% (271) & 24.2\% & 25.5\% & 28.6\% & 2.61 & 2.73\\
KG-Story  & 25.1\% (314) & 25.2\% & 25.7\% & 24.0\% & 2.49 & 2.53  \\
{\system}  & \textbf{32.6\%} (\textbf{407}) & 23.7\% & 21.7\% & \textbf{22.0\%} & \textbf{2.33} & \textbf{2.28}\\
\hline



\end{tabular}}
\end{center}
\caption{Human rankings between {\system} and three methods. The first
four columns indicate the percentage of worker rankings for each method, and the
fifth and the last column denote the average and majority ranks (1 to 4, lower is
better). 
\system outperforms other models in average ranks (p$<$0.05,~N=250), majority ranks, and also the 
percentage of 1st-rank stories.
} 
\label{tb:human_ranking}
\end{table}

\paragraph{Automatic Evaluations}

Table~\ref{tb:auto_new} shows that
the proposed method outperforms all the baselines in BLEURT,voc-d, and MLTD.
Although n-gram-based automatic metrics are known to correlate poorly with human
judgment in VIST (see Section~\ref{sec:human-eval-method}), it is still noteworthy that {\system} results in significantly lower BLEU-4 scores.
This might be cause by the fact that
{\system} uses knowledge to enrich the story content and increase lexical diversity, but could lower the performance in n-gram matching.



\begin{table}
\begin{center}
\small
\scalebox{0.84}{
\begin{tabular}{lccccc}
\hline
Method & BLEU-4 & METEOR & BLEURT & MLTD & voc-d\\
\hline\hline
AREL & \textbf{14.4} & \textbf{35.4} & 0.52 &  22.45  & 0.53\\
GLAC & 10.7 & 33.7 & 0.71 &  32.87    & 0.67  \\
KG-Story & 9.93 & 32.2 & 0.72  & 40.52   & 0.71\\
{\system} & 7.65 & 31.6 & \textbf{1.37} &\textbf{45.79} &  \textbf{0.73}\\
\hline

\end{tabular}}
\end{center}
\caption{The first two columns show the results of classic n-gram based metrics.
The third column shows BLEURT, a BERT-based metric.
The last two columns show the lexical diversity evaluation results (MLTD and voc-d).
High lexical diversity corresponds to low scores for n-gram metrics. 
}
\label{tb:auto_new}
\end{table}

\section{Ablation Study}
Three factors contribute to {\system}'s superior performance: story
elements, knowledge graphs, and plot reworking. 
To evaluate the effectiveness of each factor in our framework, we conducted three
ablation studies using human evaluations. The evaluation results are shown
in Table~\ref{tb:ablation_test}. All three experiments 
  use the same  qualitative analysis,   
and each experiment ranks {\system} and two settings with 
certain components removed.

\paragraph{Story Elements}
{\system} is compared to two models, each of which uses only \emph{objects}
or \emph{terms} for the storyline predictor to plot storylines. 
\paragraph{Knowledge Graphs}
{\system} is compared to two models, each of which uses only 
\(\mathcal{G}_{\mathit{vist}}\) or \(\mathcal{G}_{\mathit{vg}}\) for the storyline
predictor to plot storylines.  
\paragraph{Plot and Rework}
{\system} is compared to two models: one without reworking and
one without plotting or reworking. Without-reworking means the discriminator is removed,
that is, the story generator uses Equation~\ref{story_equation} for
all epochs. Without-plotting-reworking means that the storyline
predictor is additionally removed, so no frames are included;  
\emph{terms} are used directly as the story
generator's input.

\begin{figure}[t]
    \centering
    \includegraphics[width=\linewidth]{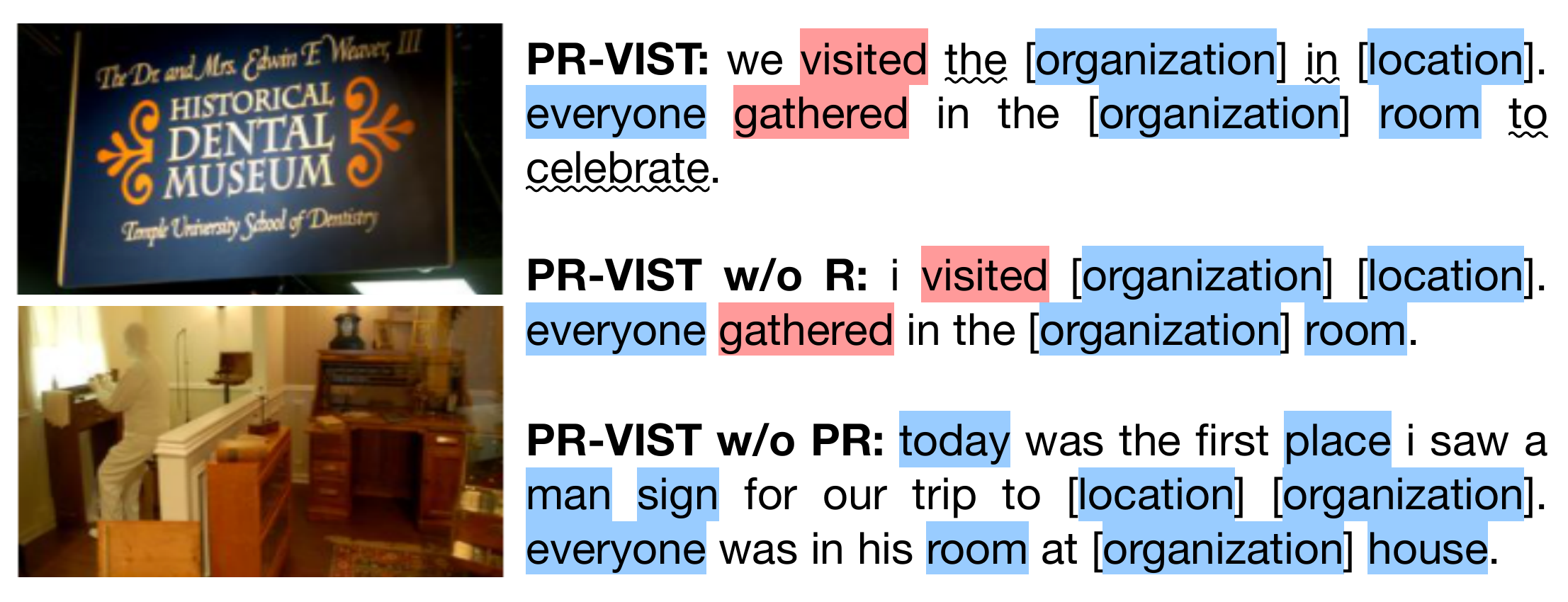}
	 \caption{Snippet of stories generated by the proposed method and two
	 configurations:
	 without reworking (R) and without plotting or reworking (PR). Nouns
	 and verb frames are denoted in blue and red. Reworked parts
	 are marked with wavy underlines.} 
    \label{fig:without}
\end{figure}

Table~\ref{tb:ablation_test} shows that {\system} outperforms all the ablation
models. Furthermore, the first and second experiments show that MTurkers prefer
story-like storylines to image-like storylines. That is, \emph{terms} and
\(\mathcal{G}_{\mathit{vist}}\) are better than \emph{objects} and
\(\mathcal{G}_{\mathit{vg}}\). For the third experiment, we note a steady
improvement from without-plotting-reworking to {\system}, showing the effectiveness 
of the proposed method. An example is shown in Figure~\ref{fig:without}.
The model cannot manage the abundant story elements without the guidance of
story plotting.
Comparing {\system} with {\system} w/o R, we see that reworking revises and
enlivens (e.g., ``[organization] in [location]'') the stories. 

\begin{table}
\begin{center}
\small
\scalebox{0.8}{

\begin{tabular}{lcccccccc}
\hline
  & \emph{objects} & \emph{terms} & \(\mathcal{G}_{\mathit{vg}}\) & \(\mathcal{G}_{\mathit{vist}}\) & Plot & Rework & Avg & Major\\
\hline\hline
&
\checkmark & \checkmark & \checkmark & \checkmark &  \checkmark &  \checkmark & 1.89 & 1.87\\
1&
 & \checkmark & \checkmark & \checkmark &  \checkmark & \checkmark  & 1.98 & 2.00\\
 &
\checkmark &  & \checkmark & \checkmark & \checkmark  &  \checkmark & 2.12 & 2.13\\
\hline
&
\checkmark & \checkmark & \checkmark & \checkmark &  \checkmark &  \checkmark & 1.97 & 1.94\\
2&
\checkmark & \checkmark &  & \checkmark &  \checkmark & \checkmark  &1.98 & 1.99\\
&
\checkmark & \checkmark & \checkmark &  & \checkmark  &  \checkmark & 2.00 & 2.06\\
\hline

&
\checkmark & \checkmark & \checkmark & \checkmark &  \checkmark &  \checkmark & 1.95 & 1.93\\
3 &
\checkmark & \checkmark & \checkmark & \checkmark &  \checkmark &  & 2.00 & 2.02\\
&
\checkmark & \checkmark & \checkmark & \checkmark &   &   & 2.03& 2.08\\

\hline

\end{tabular}}
\end{center}
\caption{Human evaluation results for ablation studies: 1.~Story elements 
2.~Knowledge graph 
3.~Plot and Rework. {\system} outperforms in all ablation
settings, indicating the importance of using all components.}
\label{tb:ablation_test}
\end{table}

\section{Discussion}

\begin{figure}[t]
    \centering
    \includegraphics[width=\linewidth ]{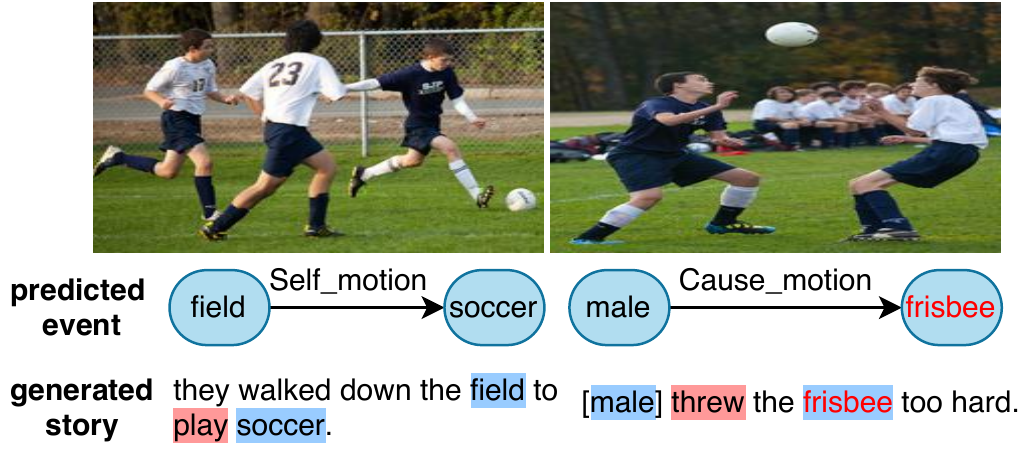}
    \caption{\system grounding error} 
    \label{fig:error}
\end{figure}

To understand areas for improvement, in the human ranking
evaluation, we asked crowd workers to select the aspect (out of six) they
disliked about the worst story (see Section~\ref{sec:human-eval-method}.)
Of the negative votes, 24.6\% were for ``grounding.''
Namely, lower-ranked stories are often not visually grounded.
We examined the outputs and found that Faster-RCNN in Stage~1 sometimes
predicts objects that are inaccurate but semantically related to the context.
Figure~\ref{fig:error} shows a typical example, where the soccer ball is
identified as a frisbee, which is incorrect but still fits the ``sports''
theme.
When the storyline predictor is unable to distinguish such mistaken objects
from appropriate objects, grounding errors occur.
A better object detector would mitigate this problem, or we could jointly optimize
plotting and generation, for instance by including reworking within storyline plotting.

\section{Conclusion}
We propose a novel story plotting and reworking framework to mimic the human
story-writing process. To the best of our knowledge, no study has integrated
knowledge graph and story plotting to utilize visual elements in VIST. Also
novel is our approximation of human-preferred stories by reusing and
aggregating story generation using the results of human-annotated story ranking
evaluations, e.g., human evaluation results from MTurk.
We also propose a novel questionnaire embedded in the comparative study to
collect detailed, meaningful human-annotated data from MTurk.
Experiments attest {\system}'s strong performance in diversity, coherence, and
humanness.

\section{Ethical Considerations}
Although our research aims to produce short stories that are vivid, engaging,
and innocent, we are aware of the possibilities of utilizing a similar approach
to generate inappropriate text (e.g., violent, racial, or
gender-insensitive stories).
The proposed visual storytelling technology enables people to generate stories
rapidly based on photo sequences at scale, which could also be used with
malicious intent, for example, to concoct fake stories using real images.
Finally, as the proposed methods use external knowledge graphs, they
reflect the issues, risks, and biases of such information sources.   
Mitigating these potential risks will require continued research.

\section{Acknowledgements}
This research is supported by Ministry of Science and Technology, Taiwan under the project contract 108-2221-E-001-012-MY3 and 108-2923-E-001-001-MY2
and the Seed Grant from the College of Information Sciences and Technology (IST), Pennsylvania State University. 
We also thank the crowd workers for participating in this project.



\bibliographystyle{acl_natbib}
\bibliography{acl2021}


\end{document}